\documentclass[10pt,twocolumn,letterpaper]{article}

\usepackage{iccv}
\usepackage{times}
\usepackage{subfigure}
\usepackage{epsfig}
\usepackage{graphicx}
\usepackage{algorithm}
\usepackage{algorithmicx}
\usepackage{algpseudocode}
\usepackage{amsmath}
\usepackage{amssymb}

\usepackage[breaklinks=true,bookmarks=false]{hyperref}

\iccvfinalcopy % *** Uncomment this line for the final submission

\def\iccvPaperID{14} % *** Enter the ICCV Paper ID here

\begin{document}

%%%%%%%%% TITLE
\title{Falls Prediction Based on Body Keypoints and Seq2Seq Architecture}

%\author{Minjie Hua \quad Yibing Nan \quad Shiguo Lian\\
%CloudMinds Technologies Inc., Beijing, China\\
%% Institution1 address\\
%{\tt\small \{michael.hua, charlie.nan, scott.lian\}@cloudminds.com}

%\and
%Yibing Nan$^*$\\
%CloudMinds Technologies\\
%% First line of institution2 address\\
%{\tt\small charlie.nan@cloudminds.com}
%\and
%Shiguo Lian\\
%CloudMinds Technologies\\
%{\tt\small scott.lian@cloudminds.com}
%}

\author{Minjie Hua\\
CloudMinds Technologies\\
{\tt\small michael.hua@cloudminds.com}
\and
Yibing Nan\\
CloudMinds Technologies\\
{\tt\small charlie.nan@cloudminds.com}
\and
Shiguo Lian\\
CloudMinds Technologies\\
{\tt\small sg\_lian@163.com}
}

\maketitle
%\thispagestyle{empty}

%%%%%%%%% ABSTRACT
\begin{abstract}
  This paper presents a novel approach for predicting the falls of people in advance from monocular video. First, all persons in the observed frames are detected and tracked with the coordinates of their body keypoints being extracted meanwhile. A keypoints vectorization method is exploited to eliminate irrelevant information in the initial coordinate representation. Then, the observed keypoint sequence of each person is input to the pose prediction module adapted from sequence-to-sequence(seq2seq) architecture to predict the future keypoint sequence. Finally, the predicted pose is analyzed by the falls classifier to judge whether the person will fall down in the future. The pose prediction module and falls classifier are trained separately and tuned jointly using Le2i dataset, which contains 191 videos of various normal daily activities as well as falls performed by several actors. The contrast experiments with mainstream raw RGB-based models show the accuracy improvement of utilizing body keypoints in falls classification. Moreover, the precognition of falls is proved effective by comparisons between models that with and without the pose prediction module.
\end{abstract}

%%%%%%%%% BODY TEXT
\section{Introduction}\label{sec:intro}

Falls are a major cause of fatal injury especially for the elderly and create a serious obstruction for independent living~\cite{FALL}. Therefore, falls prediction is one of the most meaningful applications for elderly caring and home monitoring \etc. The precognition of falls is the prerequisite for follow-up preventions and early warning, which will largely decrease the risks of falling accident. Although mainstream human action recognition (HAR) algorithms can be trained to recognize falls as one of the action classes, most of them utilize raw RGB information for classification and are not capable of predicting falls in advance.

However, the following characteristics make fall distinct from other actions: 1) Fall is highly relevant to the status of body keypoints, \ie, the body skeleton of a fallen person is significantly different. 2) Unlike smoking and handshaking, fall does not involve interactions with objects or other people. 3) Fall is an accident rather than a daily activity. So it's expected to `foresee' its happening and alert emergency as soon as possible.
%\begin{enumerate}
%  \item Falls is highly relevant to the status of body keypoints, \ie, the body skeleton of a fallen person is obviously different from others.
%  \item Unlike smoking and handshaking, falls is generally not involved in the interaction with objects or other people.
%  \item Falls is not a normal action, but an accident. So it's expected to predict its happening in advance and alert emergency as soon as possible.
%\end{enumerate}

According to 1) and 2), our falls classifier gives prediction using human body keypoints instead of raw RGB information in the video frames. The essence is to decrease the dimensionality of features with valuable cues preserved. With regard to 3), we introduce a pose prediction module adapted from sequence-to-sequence (seq2seq)~\cite{SEQ2SEQ} to predict future keypoint sequence based on the observed keypoint sequence. By analyzing future pose, the falls classifier is able to give early prediction.

It is well known that mainstream keypoints detection models like OpenPose~\cite{OPENPOSE} and AlphaPose~\cite{ALPHAPOSE} represent each extracted keypoint by its coordinate in the image. However, coordinate representation involves the body's absolute position and scale, which contribute little to action classification. Consequently, we propose a keypoints vectorization method to transform coordinates to a feature vector, in which only the direction information remains.

Since the pose prediction module already encodes temporal features, the falls classifier can focus on the spatial cues in the predicted body pose. To train the falls classifier, we re-annotated Le2i dataset~\cite{LE2I} to tag each frame a label. This operation largely increases the amount of training data, which makes the falls classifier converge better.

%\begin{figure} [tbp]
%  \centering
%  \subfigure[]{
%    \centering
%    \label{fig:case:a} %% label for first subfigure
%    \begin{minipage}[b]{0.15\textwidth}%\textwidth %0.46  %\linewidth
%      \centering
%      \includegraphics[width=1\linewidth]{figs/fig_1_a.png}%0.2
%    \end{minipage}}
%  \subfigure[]{
%    \centering
%    \label{fig:case:b}
%    \begin{minipage}[b]{0.15\textwidth}
%      \centering
%      \includegraphics[width=1\linewidth]{figs/fig_1_b.png}
%    \end{minipage}}
%  \subfigure[]{
%    \centering
%    \label{fig:case:c} %% label for first subfigure
%    \begin{minipage}[b]{0.15\textwidth}%\textwidth %0.46  %\linewidth
%      \centering
%      \includegraphics[width=1\linewidth]{figs/fig_1_c.png}%0.2
%    \end{minipage}}
%  \caption{Some cases with small obstacles in autonomous driving (a), patrol robot (b) and blind guidance (c).}
%  \label{fig:case}
%\end{figure}

The main contributions of this paper are as follows:
\begin{itemize}
  \item We proposed an end-to-end falls prediction model, which consists of a seq2seq-based pose prediction module and a falls classifier.
  \item We developed a keypoints vectorization method to extract salient features from coordinate representation.
  \item We evaluated our model by comparisons with mainstream HAR networks, and self-comparison between the pose prediction module enabled and disabled.
\end{itemize}

The rest of the paper is organized as follows: Section~\ref{sec:relat} reviews the related work on falls detection, action recognition and prediction. The proposed network is presented in Section~\ref{sec:sys}. In Section~\ref{sec:exp}, the dataset and experiments are described. Finally, Section~\ref{sec:concl} gives a conclusion.

\section{Related Work}\label{sec:relat}

\subsection{Falls Detection}

Many early solutions for falls detection relied on the wearable devices equipped by the person to be monitored. Bourke \etal proposed a method in~\cite{BO07} to detect falls based on peak acceleration. Narayanan \etal developed a distributed falls management system, which is capable of real-time falls detection using a waist-mounted triaxial accelerometer~\cite{NM07}. Bianchi \etal enhanced previous falls detection system with an extra barometric pressure sensor in ~\cite{BF09,BF10} and found that the algorithm incorporating pressure information achieved higher accuracy.

Although wearable device based falls detection methods are computationally efficient and insensitive to the environment, they require users to wear corresponding sensors for data collection. This limitation causes huge inconvenience to the users, and also makes it expensive for widespread application. Moreover, these algorithms suffer from a high false-detection rate due to sudden acceleration change in daily activities such as squat and run \etc.

Later, vision-based solutions were developed to bring more user-friendly experience. Ma \etal~\cite{MX14} proposed an approach that extracted curvature scale space (CSS) features of human silhouettes from a depth camera and represented each action by a bag of CSS words (BoCSS), which was then classified by the extreme learning machine (ELM) to identify falls. Stone \etal presented solutions for falls recognition using gait parameters in~\cite{SS11} and five handcrafted features in~\cite{SS14}. And both were based on the foreground extracted from the Kinect. Besides, Quero \etal adopted non-invasive thermal vision sensors in~\cite{QJ18} to detect falls using thermal images.

To the best of our knowledge, there was a lack of monocular vision based algorithms specialized for falls detection. However, most of the HAR models would support falls recognition after being trained on the dataset including fall as one of the actions. For example, the UCF-101 dataset~\cite{UCF101} contains 13320 videos divided into 101 action categories but without falls. By contrast, the HMDB-51~\cite{HMDB}, another large-scale dataset containing 6849 videos, takes `fall on the floor' as one of the 51 action classes.

\subsection{Action Recognition and Prediction}

As investigated in~\cite{SURVEY}, convolutional neural networks (CNN) and temporal modeling are the two major variables for action recognition. Karpathy \etal \cite{HAR2D} pioneered to introduce CNN in HAR problem by finetuning a general image recognition model pre-trained on ImageNet~\cite{IMAGENET} using UCF-101~\cite{UCF101}. However, The inability of utilizing temporal information became a severe disadvantage of 2D CNN. This issue was resolved by the 3D CNN proposed in~\cite{C3D,Conv3D}. By performing 3D convolutions, the 3D CNN was capable of extracting features from both spatial and temporal dimensions, thereby capturing the motion information in multiple adjacent frames. As for the temporal modeling methods like Two-Stream~\cite{TWOSTREAM} and TSN~\cite{TSN}, the main idea was to extract spatial features using 2D convolutions and encode temporal information by recurrent neural networks(RNN).

\begin{figure}[htbp]%
\centering
\includegraphics[width=1\linewidth]{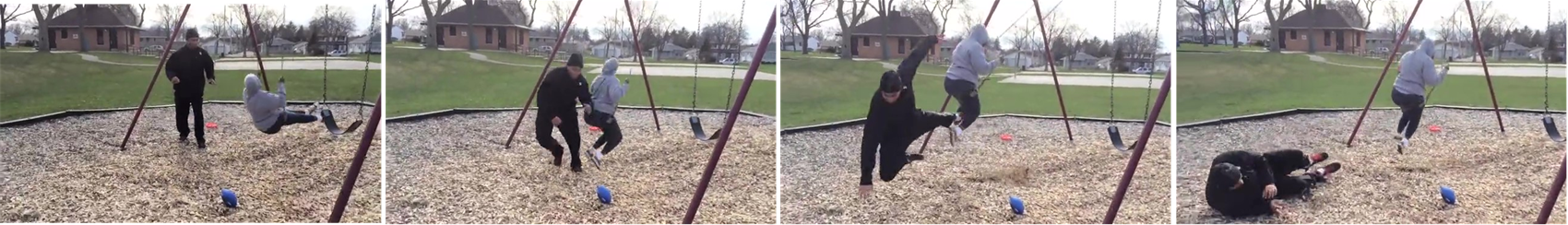}
\caption{An example of multiple actions presented in the meantime. TSN model trained on UCF-101 dataset gives the prediction `swing'. However, the man's fall is ignored because only one label will be predicted for a video.}
\label{fig:multiact}
\end{figure}

However, all the aforementioned models predict only one label for each video, even when multiple actions are existing in the meantime. As shown in Fig.~\ref{fig:multiact}, although two people are presenting different actions, the TSN model~\cite{TSN} trained on UCF-101 dataset~\cite{UCF101} just predict `swing' with the man's fall being ignored.

Action recognition models must `watch' the entire video to give prediction. But in some cases like incomplete data or the requirement of early alarm, predicting the future action based on partial clip is necessary. Early classification and motion prediction are two routes of action prediction towards different goals: Given $t_{\text{obs}}$ observed frames $(f_1, f_2, ..., f_{t_{\text{obs}}})$, early classification tries to infer the label $y$ in advance, while motion prediction tries to produce future motions in next $t_{\text{pred}}$ frames $(f_{t_{\text{obs}}+1}, f_{t_{\text{obs}}+2}, ..., f_{t_{\text{obs}}+t_{\text{pred}}})$.

The work in~\cite{MS16} designed novel ranking losses to learn activity progressing in LSTMs for early classification. Kong \etal adopted an auto-encoder to reconstruct missing features from observed frames by learning from the complete action videos~\cite{KT17}. In \cite{KG18}, a mem-LSTM model was proposed to store several hard-to-predict samples and a variety of early observations in order to improve the prediction performance at early stage.

\begin{figure*}[tbp]%
\centering
\includegraphics[width=1\linewidth]{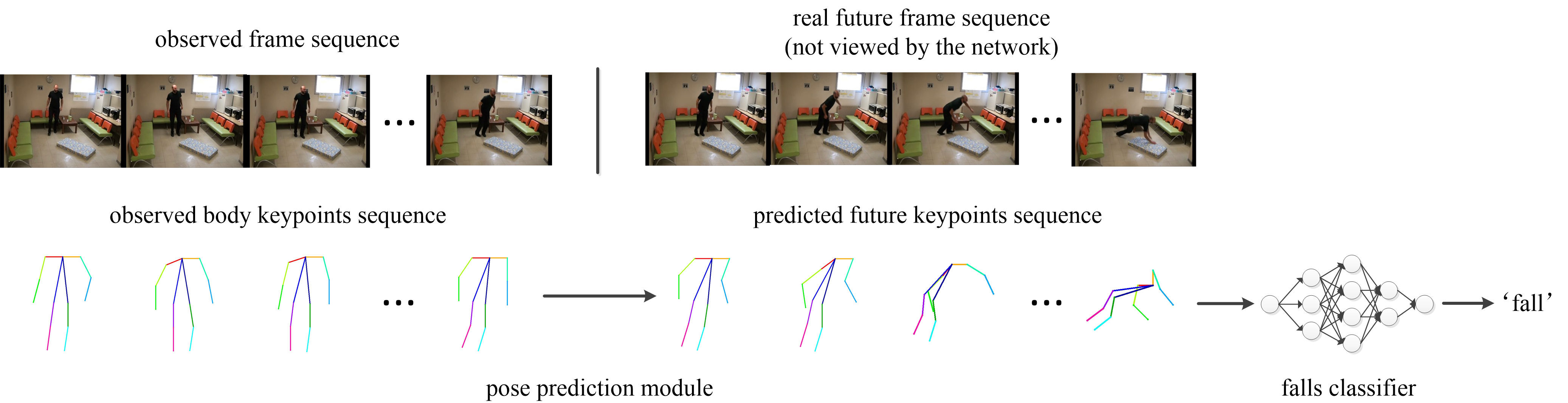}
\caption{The major work flow of our model. A sequence of observed frames is input to the network. Then the human body keypoints are extracted from each frame to form a keypoints sequence, which is used to predict future keypoints sequence by the pose prediction module. At last, the predicted body pose is passed into a falls classifier to judge whether the person will fall down in the future.}
\label{fig:workflow}
\end{figure*}

In recent years, motion prediction attracted much attention. Fragkiadaki \etal proposed the encoder-recurrent-decoder (ERD) model~\cite{ERD} that extended long short term memory (LSTM)~\cite{LSTM} to jointly learn representations and their dynamics. Jain \etal proposed structural-RNN (SRNN) based on spatiotemporal graphs (st-graphs) in \cite{SRNN} to capture the interactions between the human and the objects. Martinez \etal modified standard RNN models in sampling-based loss and residual architectures for better motion prediction~\cite{CVPR17}. Tang \etal proposed modified highway unit (MHU) to filter the still keypoints and adopted gram matrix loss in~\cite{IJCAI18} for long-term motion prediction. For the same purpose, Li \etal proposed convolutional encoding module (CEM) in~\cite{CVPR18} to learn the correlations between each keypoint, which is hard for an RNN model.

\section{Methodology}\label{sec:sys}

% As illustrated in Fig.~\ref{fig:flow}, the proposed obstacle detection and avoidance system contains several steps: RGB-D based two-stage semantic segmentation, morphological processing, local destination setting and path planning. The two-stage semantic segmentation transforms the input RGB-D image to a raw binary image, which is then smoothed in morphological processing. As a result, the module generates a calibrated binary image where every pixel is labelled as either ground or obstacle. Then the binary image is passed to the obstacle avoidance module to determine a destination and a walkable path. The whole process works repeatedly during robot¡¯s moving.

\subsection{Overview of the Proposed Model}

The problem to be solved in this paper is formulated as follow: Given $t_{\text{obs}}$ observed frames $(f_1, f_2, ..., f_{t_{\text{obs}}})$, we try to predict whether the human(s) in the video will fall down in next $t_{\text{pred}}$ frames (\ie, from $f_{t_{\text{obs}} + 1}$ to $f_{t_{\text{obs}} + t_{\text{pred}}}$).

The skeleton framework of our model is presented in Fig.~\ref{fig:workflow}. The input is a sequence of observed frames. We first adopted OpenPose~\cite{OPENPOSE} to extract keypoints coordinates of human(s) from each observed frame. The bounding boxes of detected persons were passed to DeepSort~\cite{DEEPSORT}, a tracking algorithm, to cluster body keypoints belonging to the same person in different frames. As a result, the $i$-th person was corresponding to a sequence of observed keypoints $\mathbf{K}^i_{obs}=\left(\mathbf{k}^i_1, \mathbf{k}^i_2, ..., \mathbf{k}^i_{t_{\text{obs}}} \right)$, where $\mathbf{k}^i_j$ included the keypoints coordinates of the $i$-th person in frame $j$.

Based on the observation that fall is highly correlated to the relative position between body keypoints, we exploited a keypoints vectorization method to extract salient features from coordinate representation. The transformed sequence of the $i$-th person was denoted $\overline{\mathbf{K}^i_{obs}}=\left(\overline{\mathbf{k}^i_1}, \overline{\mathbf{k}^i_2}, ..., \overline{\mathbf{k}^i_{t_{\text{obs}}}} \right)$. Then, the pose prediction module adapted from seq2seq architecture~\cite{SEQ2SEQ} was used to predict body poses in next $t_{\text{pred}}$ frames. Considering that applying excessive LSTM units made the network hard to converge, we encoded several consecutive keypoints vectors in one LSTM unit to shorten the lengths of both encoder and decoder LSTM layers. Moreover, shorter sequences also suppressed the mean-pose problem caused by long-term prediction~\cite{IJCAI18,CVPR18}.

After that, $\overline{\mathbf{k}^i_{t_{\text{obs}}+t_{\text{pred}}}}$, the future keypoints vector of the $i$-th person at frame $f_{t_{\text{obs}} + t_{\text{pred}}}$, was predicted and used for classification. The falls classifier adopted fully connected network and was trained on re-annotated Le2i~\cite{LE2I} dataset, in which each frame was labeled either `fall' or `no fall'. Combining the pose prediction module and falls classifier, our model was capable of predicting falls in advance.

\subsection{Keypoints Vectorization}\label{subsec:vector}

The keypoints coordinates extracted by OpenPose~\cite{OPENPOSE} cannot reflect the correlation between different keypoints, and suffered from the effects of body skeleton's absolute position and scale. With the motivation that the same pose should be represented by the same keypoints vector, we exploited the following keypoints vectorization method.

\begin{figure}[tbp]%
\centering
\includegraphics[width=0.7\linewidth]{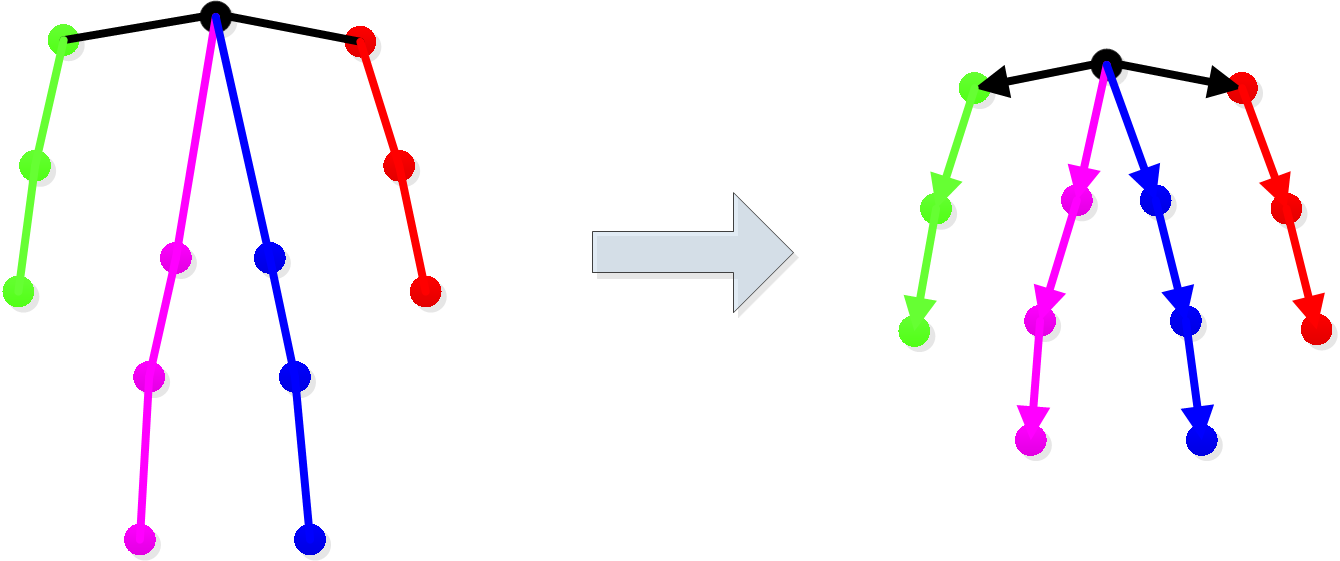}
\caption{The illustration of the keypoints vectorization method. The arrows indicate 12 vectors constructed from the coordinates of 13 body keypoints. All the vectors are normalized to unit length with only direction information remains.}
\label{fig:kptvec}
\end{figure}

As we know, the 18 keypoints of MS COCO~\cite{MSCOCO} are nose, neck, left and right shoulders, elbows, wrists, hips, knees, ankles, eyes and ears. Since we focused on the body keypoints, 5 face keypoints (\ie nose, left and right eyes and ears) were ignored. For 13 concerned ones, we transformed their coordinates to vectors connecting with corresponding adjacent keypoints. As illustrated in Fig.~\ref{fig:kptvec}, the left and right shoulders are connected to the neck, the left/right elbow is connected to left/right shoulder, and the left/right wrist is connected to left/right elbow. Similarly, the left and right hips are connected to the neck, the left/right knee is connected to left/right hip, and the left/right ankle is connected to left/right knee. As a result, 12 keypoints vectors were constructed from the coordinates information. Finally, we normalized all the vectors to unit length.

Formally, recall that the observed keypoints sequence of the $i$-th person was $\mathbf{K}^i_{obs}=\left(\mathbf{k}^i_1, \mathbf{k}^i_2, ..., \mathbf{k}^i_{t_{\text{obs}}} \right)$, where
\begin{equation}\label{eq:transk}
\begin{aligned}
  \mathbf{k}^i_j=\left(x^i_{j,1}, y^i_{j,1}, x^i_{j,2}, y^i_{j,2}, ..., x^i_{j,18}, y^i_{j,18} \right).
\end{aligned}
\end{equation}
$(x^i_{j,m}, y^i_{j,m})$ denoted the $m$-th keypoint coordinate of the $i$-th person in frame $j$. For the $p$-th connection pointing from the $l$-th keypoint to the $r$-th keypoint, the keypoints vector $\left(\overline{x^i_{j,p}}, \overline{y^i_{j,p}} \right)$ was calculated by:
\begin{equation}\label{eq:transfv}
\begin{aligned}
  \left( {\overline {x_{j,p}^i} ,\overline {y_{j,p}^i} } \right) = {\textstyle{{\left( {x_{j,r}^i - x_{j,l}^i,y_{j,r}^i - y_{j,l}^i} \right)} \over {\sqrt {{{\left( {x_{j,r}^i - x_{j,l}^i} \right)}^2} + {{\left( {y_{j,r}^i - y_{j,l}^i} \right)}^2}} }}}.
\end{aligned}
\end{equation}
After 12 keypoints vectors were constructed, they were then concatenated to form  $\overline{\mathbf{k}^i_j}$ as follows:
\begin{equation}\label{eq:transkv}
\begin{aligned}
  \overline{\mathbf{k}^i_j}=\left(\overline{x^i_{j,1}}, \overline{y^i_{j,1}}, \overline{x^i_{j,2}}, \overline{y^i_{j,2}}, ..., \overline{x^i_{j,12}}, \overline{y^i_{j,12}} \right).
\end{aligned}
\end{equation}

The transformation from $\mathbf{k}^i_j$ to $\overline{\mathbf{k}^i_j}$ eliminates the absolute position and scale of body skeleton, and preserves direction information between adjacent keypoints. It not only ensures that the same body pose has the same representation, but also extracts salient features for better falls classification.

\subsection{Seq2Seq-based Pose Prediction Module}\label{subsec:ppm}

The authors in~\cite{SEQ2SEQ} proposed a seq2seq network, which was applied to machine translation at first and achieved excellent performance. Later, they introduced this architecture to conversational modeling in~\cite{CONVER}. In analogy to mapping a sentence from one language to another in machine translation, the conversational model maps a query sentence to a response sentence. Generally, the seq2seq framework uses an LSTM~\cite{LSTM} layer to encode the input sentence to a vector of a fixed dimensionality, and then another LSTM layer to decode the target sentence from the vector. This encoder-decoder architecture is widely used in sequence mapping problems such as machine translation~\cite{SEQ2SEQ}, conversation modeling~\cite{CONVER} and even video caption~\cite{VID2TEXT} because of its powerful capabilities.

Inspired by the great success of seq2seq network in the sequence mapping problems, we implemented a seq2seq-based pose prediction module to predict body poses in the future $t_{\text{pred}}$ frames. Formally, recall that the observed keypoints vector sequence of the $i$-th person was $\overline{\mathbf{K}^i_{obs}}=\left(\overline{\mathbf{k}^i_1}, \overline{\mathbf{k}^i_2}, ..., \overline{\mathbf{k}^i_{t_{\text{obs}}}} \right)$. The pose prediction module was designed to generate future keypoints vector sequence $\overline{\mathbf{K}^i_{pred}}=\left(\overline{\mathbf{k}^i_{t_{\text{obs}} + 1}}, \overline{\mathbf{k}^i_{t_{\text{obs}} + 2}}, ..., \overline{\mathbf{k}^i_{t_{\text{obs}} + t_{\text{pred}}}} \right)$.

\begin{figure*}[htbp]%
\centering
\includegraphics[width=1\linewidth]{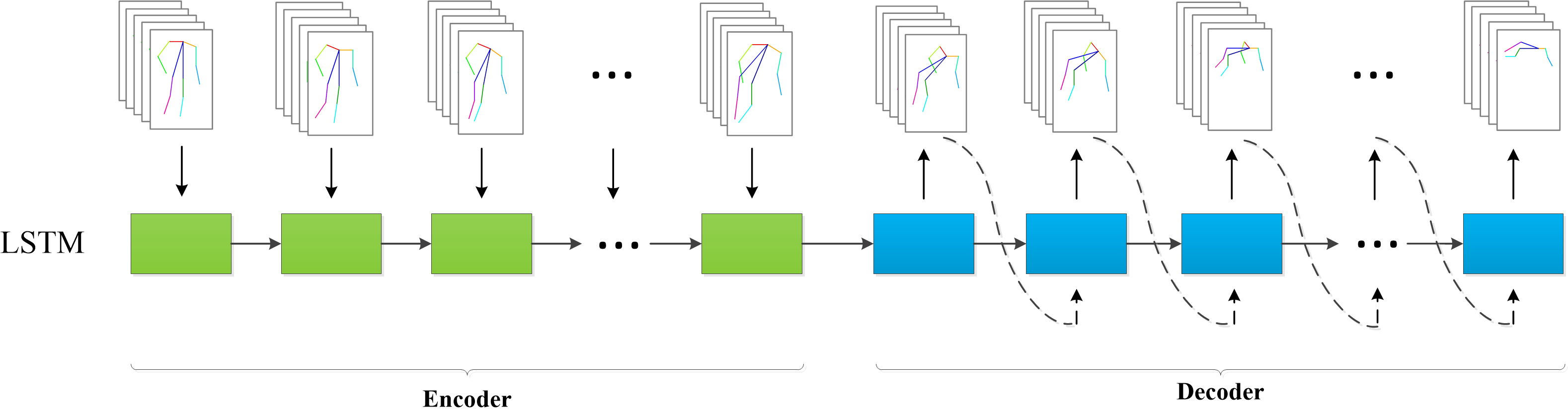}
\caption{The architecture of our seq2seq-based pose prediction module, which is composed of an encoder (colored green), and a decoder (colored blue). Each LSTM unit in the encoder parses an observed keypoints vector (visualized in the figure to present more intuitive result) and produces a hidden vector. The first unit in the decoder accepts the last hidden vector from the encoder and generates the first prediction. Latter units receive the previous prediction and produce a new one. Note that the vector packing is reflected in the figure.}
\label{fig:seqarch}
\end{figure*}

As illustrated in Fig.~\ref{fig:seqarch}, the pose prediction module is composed of two LSTM~\cite{LSTM} layers as the encoder and decoder respectively. The encoder analyzes the sequence of observed keypoints vectors with each LSTM unit parsing one keypoints vector. The hidden vector calculated by the previous unit is passed to the next one. In the decoder, the keypoints vector is generated one at a step. The first decoder LSTM unit accepts the last hidden vector from the encoder and outputs the first prediction. Latter units take the previous outcome as input and produce a new one. As all the LSTM units in the decoder complete predictions, the output keypoints sequence will be generated.

Recall the mechanism of LSTM layer: Assume that the input sequence is $\left( {{\mathbf{x}}_{1}},{{\mathbf{x}}_{2}},...,{{\mathbf{x}}_{m}} \right)$, the $t$-th LSTM unit updates the states based on the states at $t-1$:
\begin{eqnarray}\label{eq:state}
\begin{aligned}
& {{\mathbf{i}}_{t}} = \sigma \left( {{\mathbf{W}}_{i}}\left[ {{\mathbf{h}}_{t-1}},{{\mathbf{x}}_{t}} \right]+{{\mathbf{b}}_{i}} \right) \\
& {{\mathbf{f}}_{t}} = \sigma \left( {{\mathbf{W}}_{f}}\left[ {{\mathbf{h}}_{t-1}},{{\mathbf{x}}_{t}} \right]+{{\mathbf{b}}_{f}} \right) \\
& {{\mathbf{o}}_{t}} = \sigma \left( {{\mathbf{W}}_{o}}\left[ {{\mathbf{h}}_{t-1}},{{\mathbf{x}}_{t}} \right]+{{\mathbf{b}}_{o}} \right) \\
& \widetilde{{{\mathbf{c}}_{t}}} = \tanh \left( {{\mathbf{W}}_{c}}\left[ {{\mathbf{h}}_{t-1}},{{\mathbf{x}}_{t}} \right]+{{\mathbf{b}}_{c}} \right) \\
& {{\mathbf{c}}_{t}} = {{\mathbf{f}}_{t}}\odot {{\mathbf{c}}_{t-1}}+{{\mathbf{i}}_{t}}\odot \widetilde{{{\mathbf{c}}_{t}}} \\
& {{\mathbf{h}}_{t}} = {{\mathbf{o}}_{t}}\odot \tanh \left( {{\mathbf{c}}_{t}} \right),
\end{aligned}
\end{eqnarray}
where $\sigma$ denotes the sigmoid function, $\tanh$ is the hyperbolic tangent function, $\odot$ denotes the element-wise multiplication, ${{\mathbf{i}}_{t}}$, ${{\mathbf{f}}_{t}}$, ${{\mathbf{o}}_{t}}$, ${{\mathbf{c}}_{t}}$ and ${{\mathbf{h}}_{t}}$ represent input gate, forget gate, output gate, cell state and hidden state of the $t$-th LSTM unit respectively. $\mathbf{W}$ and $\mathbf{b}$ are trainable weights.

We attempted to apply $t_{\text{obs}}$ LSTM units to the encoder and $t_{\text{pred}}$ LSTM units to the decoder (\ie, each LSTM unit only deals with a single keypoints vector). However, we found that as the increment of $t_{\text{obs}}$ and $t_{\text{pred}}$, the network was getting harder to converge. To mitigate the negative effect of long LSTM layer, we packed every $n_p$ consecutive keypoints vectors in one for decreasing the length of keypoints sequence and the number of LSTM units. Formally, the packed sequence of $\overline{\mathbf{K}^i_{obs}}=\left(\overline{\mathbf{k}^i_1}, \overline{\mathbf{k}^i_2}, ..., \overline{\mathbf{k}^i_{t_{\text{obs}}}} \right)$ was $\widetilde{\mathbf{K}^i_{obs}}=\left(\widetilde{\mathbf{k}^i_1}, \widetilde{\mathbf{k}^i_2}, ..., \widetilde{\mathbf{k}^i_{t_{\text{obs}}/n_p}} \right)$, where
\begin{equation}\label{eq:concat}
\begin{aligned}
  \widetilde{\mathbf{k}^i_j}=\overline{\mathbf{k}^i_{n_p(j-1)+1}}\oplus \overline{\mathbf{k}^i_{n_p(j-1)+2}}\oplus ... \oplus \overline{\mathbf{k}^i_{n_pj}},
\end{aligned}
\end{equation}
where $\oplus$ denotes the concatenation of two vectors. If there are no enough keypoints vectors for the last package (it will always happen when $n_p$ is not divisible by $t_{\text{obs}}$), zero-paddings are filled to its tail for the dimensional equality. Correspondingly, since the vectors in output sequence are also packed, they need to be unpacked before classification to obtain the future pose at each frame.

Benefiting from the vector packing technique, the pose prediction module required less time to get convergent in the training phase, and the mean-pose problem raised in~\cite{IJCAI18,CVPR18} was also suppressed.

\subsection{Falls Classifier}\label{subsec:cls}

The falls classifier was trained to perform inference on the future keypoints vector at frame $f_{t_{\text{obs}} + t_{\text{pred}}}$. Considering that $\overline{\mathbf{k}^i_{t_{\text{obs}}+t_{\text{pred}}}}$ only contained 24 features, we simply adopted a traditional fully connected neural network for classification. The input layer was embedded with 24 neurons to fit the dimensionality of the input vector. And we set up five hidden layers containing 96, 192, 192, 96, 24 neurons respectively. The output layer with 2 neurons was used to give the final prediction: `fall' or `no fall'.

%\begin{equation}\label{eq:a2}
%\begin{aligned}
%  {{a}_{2}}=f({{k}_{2}}\cdot \min (w,h))
%\end{aligned}
%\end{equation}
%\begin{equation}\label{eq:a3}
%\begin{aligned}
%  {{a}_{3}}=f({{k}_{3}}\cdot \min (w,h))
%\end{aligned}
%\end{equation}

%\begin{figure}[htbp]%
%\centering
%\includegraphics[width=1\linewidth]{figs/fig_5.png}
%\caption{Morphological processing on binarized semantic segmentation result.}
%\label{fig:morlo}
%\end{figure}
%\begin{figure} [htbp]
%  \centering
%  \subfigure[]{
%    \centering
%    \label{fig:morlo:a} %% label for first subfigure
%    \begin{minipage}[b]{0.23\textwidth}%\textwidth %0.46  %\linewidth
%      \centering
%      \includegraphics[width=1\linewidth]{figs/fig_5_a.png}%0.2
%    \end{minipage}}
%  \subfigure[]{
%    \centering
%    \label{fig:morlo:b}
%    \begin{minipage}[b]{0.23\textwidth}
%      \centering
%      \includegraphics[width=1\linewidth]{figs/fig_5_b.png}
%    \end{minipage}}
%  \caption{Morphological processing results from binarized semantic segmentation maps. (a) indoor scenario. (b) outdoor scenario.}
%  \label{fig:morlo}
%\end{figure}

\section{Experiments}\label{sec:exp}

\subsection{Dataset Overview}

We trained and evaluated our model on Le2i falls detection dataset~\cite{LE2I}, which consists of 191 videos captured under four different scenes: home, coffee room, office and lecture room. The frame rate is 25 frames per second and the resolution is 320$\times$240 pixels. In each video, an actor performs various of normal activities and falls (fall might be absent in several videos). The official annotations provide the falling-start frame stamp and the falling-end frame stamp for each video. If there is no fall in a video, both frame stamps will be marked as 0 in its annotation file.

For the requirement of tagging a label for each frame, we first looked through all the videos and manually annotated an extra getting-up frame stamp for each one. If there is no fall appears, this value will be set to 0. And if the actor does not get up until the video ends, this value will be set to the last frame. Suppose that the frame stamps of falling start, falling end and getting up are denoted $S_{fs}$, $S_{fe}$, $S_{gu}$ respectively, we attempted three automatic frame-annotation principles in our experiments:
\begin{enumerate}
  \item Frames between $S_{fs}$ and $S_{fe}$ are labeled `fall';
  \item Frames between $S_{fs}$ and $S_{gu}$ are labeled `fall';
  \item Frames between $S_{fe}$ and $S_{gu}$ are labeled `fall'.
\end{enumerate}
And all the excluded frames are labeled `no fall'. The first principle only regarded the falling proceeding as fall, which means `no fall' will be annotated to a fallen person. Under this principle, the trained falls classifier could not recognize falls normally. The second one resulted in a high false positive rate because many `precursors' of falling event will be predicted as `fall' even when they are not leading to a real fall. The third principle achieved great balance by labeling `fall' after the actor was already in the fallen state. So we finally adopted this principle for frame annotations.

\subsection{Experiments Setup}

We implemented our model on a workstation with double Nvidia 1080Ti GPUs. The seq2seq-based pose prediction module and the falls classifier were trained separately and tuned jointly.

To train the pose prediction module, we preprocessed all videos in the Le2i dataset. For each video, we utilized OpenPose to extract the actor's keypoints coordinates frame by frame, and transformed them to keypoints vectors with the method proposed in Section~\ref{subsec:vector}. However, due to the effect of illuminance or camera perspective \etc, the actor's keypoints might be partially or completely missed. We dealt with the partial missing by setting vectors connecting undetected keypoint to $(0, 0)$ in the vectorization step. For the complete missing, the corresponding frames would be discarded. To ensure the coherence of keypoints sequence, consecutive 10 discarded frames would break the sequence into two segments. Then we filtered out the sequences containing less than 10 frames and finally obtained 139 sequences. The maximum, minimum and average frames of these sequences are 1773, 13 and 241.26 respectively.

During the training phase, the network loaded all the processed keypoints sequences and acquired training samples according to the configurations of $t_{\text{obs}}$ and $t_{\text{pred}}$. For each sequence, all sub-sequences with length of $t_{\text{obs}} + t_{\text{pred}}$ frames were segmented and used as valid training samples. The former $t_{\text{obs}}$ frames were input to the encoder of the pose prediction module, and the latter $t_{\text{pred}}$ ones were regarded as ground truth. After the network completed inference, the mean square error (MSE) was applied to calculate the loss between ground truth and the predicted sequence. And the network was optimized using Adam~\cite{ADAM} algorithm. In our experiments, the learning rate was set to 0.001.

To evaluate performance of the pose prediction module trained with different selections of $t_{\text{obs}}$, $t_{\text{pred}}$ and $n_p$ mentioned in Section~\ref{subsec:ppm}, we exploited mean cosine similarity (MCS) as the metrics. Specifically, suppose that there are $m$ test samples, the ground truth and the predicted sequence of the $i$-th sample are $\mathbf{g}^i$ and $\mathbf{p}^i$ respectively. Note that $\mathbf{g}^i$ and $\mathbf{p}^i$ both contain $t_{\text{pred}}$ keypoints vectors ($\mathbf{p}^i$ has been unpacked). The $j$-th vector of $\mathbf{g}^i$ and $\mathbf{p}^i$ are denoted $\mathbf{g}^i_j$ and $\mathbf{p}^i_j$. Then the MCS was calculated by:
\begin{equation}\label{eq:mcs}
\begin{aligned}
	{\text{MCS}} = \frac{1}{m}\sum\limits_{i = 1}^m {{{C}^i}},
\end{aligned}
\end{equation}
where
\begin{equation}\label{eq:ci}
\begin{aligned}
	{{C}^i} = \frac{1}{{{t_{{\text{pred}}}}}}\sum\limits_{j = 1}^{{t_{{\text{pred}}}}} {\frac{{{\mathbf{g}}_j^i \cdot {\mathbf{p}}_j^i}}{{\left\| {{\mathbf{g}}_j^i} \right\|\left\| {{\mathbf{p}}_j^i} \right\|}}}.
\end{aligned}
\end{equation}

\begin{figure}[tbp]%
\centering
\includegraphics[width=1\linewidth]{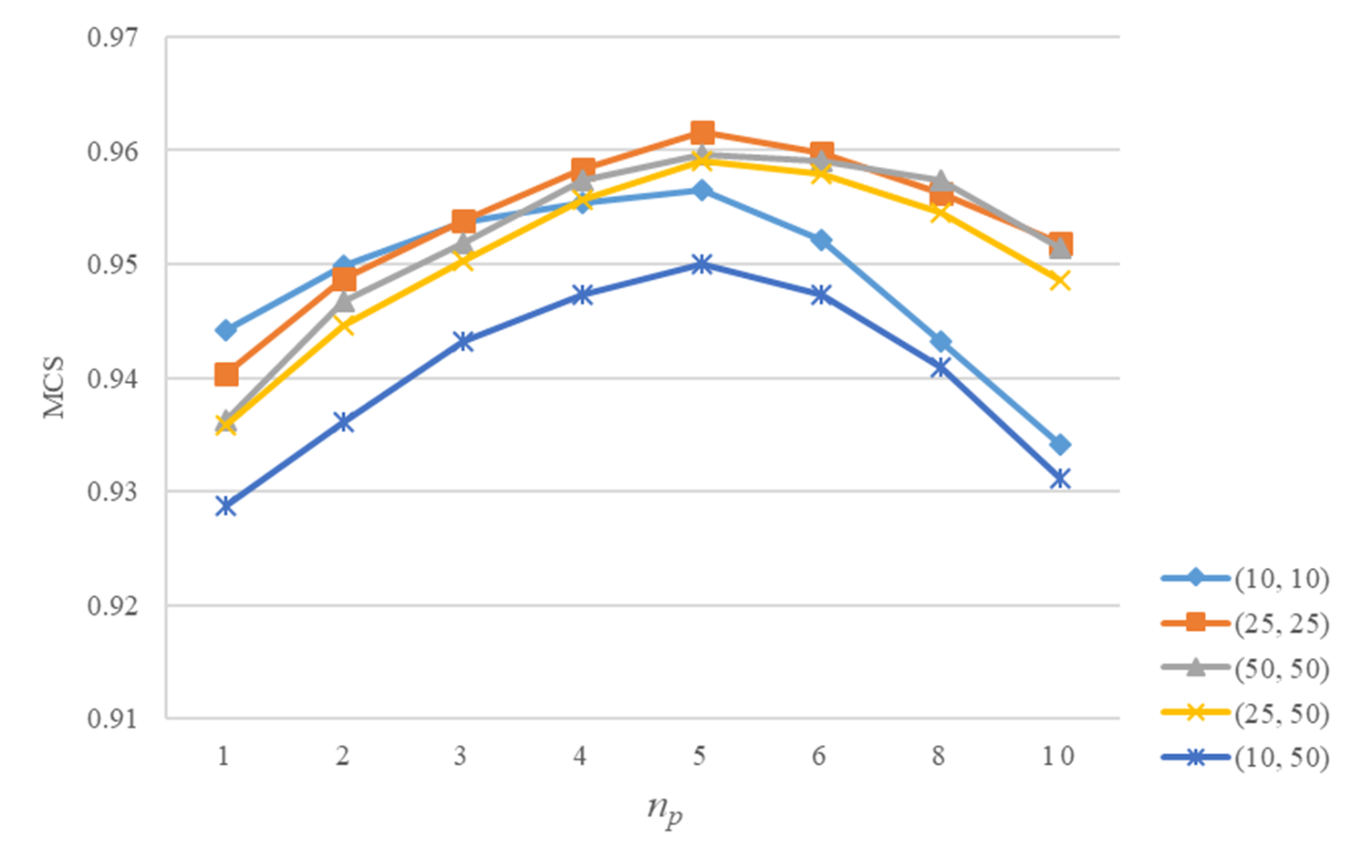}
\caption{The MCS comparisons among several parameter configurations. We first selected five combinations of $t_{\text{obs}}$ and $t_{\text{pred}}$, which is denoted by ($t_{\text{obs}}$, $t_{\text{pred}}$) in the figure, then trained and evaluated the pose prediction module with different $n_p$.}
\label{fig:seqparam}
\end{figure}

We tried several configurations of $t_{\text{obs}}$, $t_{\text{pred}}$ and $n_p$ to see the effect to the pose prediction module. It can be seen from Fig.~\ref{fig:seqparam} that observing 10 frames was insufficient for the network to give accurate predictions, especially for long predictions. Using 25 frames to predict 25 frames and using 50 frames to predict 50 frames both resulted in high MCS. Generally, the network was capable of predicting future 50 frames based on 25 observed frames with a tolerable MCS decrease. And $n_p = 5$ showed the best performance among all the candidate values. Hoping that the pose prediction module could predict further future poses with acceptable performance for earlier falls warning, we adopted $t_{\text{obs}} = 25$, $t_{\text{pred}} = 50$ and $n_p = 5$ in the deployment.

With respect to the training of falls classifier described in~\ref{subsec:cls}, we mapped each keypoints vector to the label of corresponding frame. Samples with less than 8 detected body keypoints were discarded because they contained too many null features. (In the inference phase, these samples would be prejudged as `unknown' before the classification.) All the valid vector-label pairs were used to train the network and cross-entropy function was adopted to calculate the loss. Adam~\cite{ADAM} was still selected as the optimizer.

\subsection{Evaluations}

Evaluations are designed concerning about the following questions: 1) Whether our falls classifier utilizing body keypoints information outperforms mainstream raw RGB based models. 2) What effects will the pose prediction module bring to the accuracy of falls classification.
%\begin{enumerate}
%  \item Whether our falls classifier utilizing body keypoints information outperforms raw RGB based models.
%  \item What effects will the pose prediction module bring to the accuracy of falls classification.
%\end{enumerate}

\subsubsection{Body Keypoints \vs Raw RGB}
We conducted comparisons between our falls classifier and popular raw RGB-based HAR models including C3D~\cite{C3D}, Two-Stream~\cite{TWOSTREAM} and TSN~\cite{TSN} \etc. However, considering that these models just predict one label to a complete video rather than a single frame, we converted the annotations from frame-label pairs to clip-label pairs.

The main idea was to segment 3-second (\ie 75 frames) clips from the original video and labeled `fall' to those including the whole falling proceeding and `no fall' to those not involving falls at all. The choice of 3-second length is based on the statistic that the average duration of falling proceeding among all the videos is 1.26 seconds, which ensures the acquirement of positive samples. Formally, recall that $S_{fs}$, $S_{fe}$ and $S_{gu}$ are three frame stamps that mark the falling start, falling end and getting up respectively. Suppose a clip is segmented from the video's $S_l$-th frame to $S_r$-th frame ($S_r-S_l=75$), its annotation is decided by:
\begin{equation}\label{eq:slsr}
\begin{aligned}
\begin{array}{l}
{S_l} \le {S_{fs}}\text{ and } S_{fe} \le {S_r} \le {S_{gu}}\quad\text{labeled `fall'}\\
{S_r} \le {S_{fs}}\,\text{ or }\;  {S_l} \ge {S_{gu}}\qquad\;\;\:\text{labeled `no fall'}
\end{array}
\end{aligned}
\end{equation}

To avoid ambiguities, clips with partial falling proceeding were excluded. We finally obtained 9549 samples, which were randomly divided into a training set and a test set with the ratio of 7:3. We finetuned C3D (3 nets), Two-Stream and TSN (2 modalities) on the training set and evaluated them on the test set. For each clip, our falls classifier directly classified the keypoints vector from the frame $S_r$. The limitation of ${S_r} \le {S_{gu}}$ in Eq.~\eqref{eq:slsr} ensured the label consistency between each clip and its last frame.

\begin{table}[bp]
\center
  \caption{Comparisons between mainstream RGB-based models and our keypoints-based model on falls classification problem}
  \label{tb:comprgb}
\begin{center}
\begin{tabular}{l|c|c|c|c}
\hline
\qquad Model & Acc. & Prec. & Rec. & F1\\
\hline
C3D~\cite{C3D} & 89.4\% & 66.1\% & 87.9\% & 0.755\\
Two-Stream~\cite{TWOSTREAM} & 91.6\% & 71.6\% & 91.0\% & 0.801\\
TSN~\cite{TSN} & 94.6\% & 79.8\% & 94.5\% & 0.866\\
Ours & \textbf{97.8\%} & \textbf{90.8\%} & \textbf{98.3\%} & \textbf{0.944}\\
\hline
\end{tabular}
\end{center}
\end{table}

The test set consists of 2865 samples, including 531 positive samples (fall) and 2334 negative samples (no fall). We adopted accuracy, precision, recall and F1-score to evaluate each model. As illustrated in Table~\ref{tb:comprgb}, our keypoints-based classifier showed better performance than mainstream raw RGB-based models, which proved the salience of body keypoints features for falls classification problem.

\begin{figure*}[tbp]%
\centering
\includegraphics[width=1\linewidth]{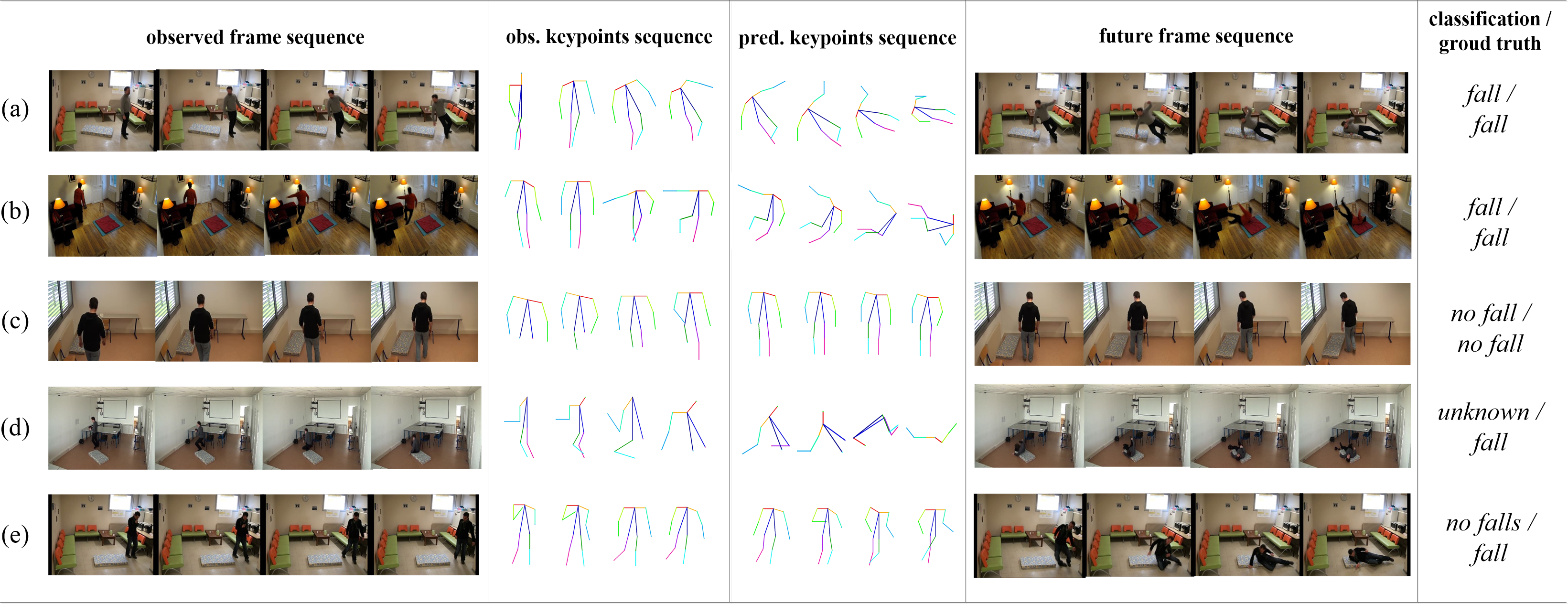}
\caption{Some experimental results of our model. In case (a)-(c), the pose prediction module successfully predicts future keypoints sequence based on the observed frame sequence, and the falls classifier correctly infers the label. (d) and (e) are two failure cases. In (d), there are too many missing detections of the body keypoints due to the camera perspective. Consequently, the pose prediction module generates absurd keypoints prediction, which is insufficient for classification. In (e), the observed keypoints sequence does not includes any `precursors' of falling event, thus leads to wrong prediction and classification.}
\label{fig:result}
\end{figure*}

\subsubsection{Effects of the Pose Prediction Module}

The pose prediction module was evaluated by self comparisons with this module enabled (denoted as ${Model}_\text{pred+cls}$) and disabled (denoted as ${Model}_\text{cls}$).

The first experiment was designed to compare the classification results between predicted keypoints and directly observed ones. Specifically, while predicting the label of the $i$-th frame in a video, ${Model}_\text{pred+cls}$ utilized frames from $i-74$ to $i-50$ (25 frames) to predict the keypoints vectors of frames from $i-49$ to $i$ (50 frames). Then the predicted keypoints vector of frame $i$ was input to falls classifier to produce a label. As for ${Model}_\text{cls}$, it was directly given the keypoints vector of the $i$-th frame for classification.

\begin{table}[tbp]
\center
  \caption{Comparison between ${Model}_\text{cls}$ and ${Model}_\text{pred+cls}$}
  \label{tb:compred1}
\begin{center}
\begin{tabular}{l|c|c|c|c}
\hline
\qquad Model & Acc. & Prec. & Rec. & F1\\
\hline
${Model}_\text{cls}$ & \textbf{99.2\%} & \textbf{95.1\%} & \textbf{98.8\%} & \textbf{0.970}\\
${Model}_\text{pred+cls}$ & 98.7\% & 92.6\% & 98.0\% & 0.952\\
\hline
\end{tabular}
\end{center}
\end{table}

The comparison result is presented in Table~\ref{tb:compred1}. In accordance with expectations, the pose prediction module brought a reduction to the accuracy of falls classification.

However, ${Model}_\text{pred+cls}$ is capable of predicting falls in advance. For a fairer comparison, we conducted another experiment to enable ${Model}_\text{cls}$ to `foresee' the falls by early annotation. Specifically, the labels of the falling proceeding (\ie, from frame $S_{fs}$ to $S_{fe}$) were modified to `fall' in the training of ${Model}_\text{cls}$. As shown in Table~\ref{tb:compred2}, compared to the pose prediction method, early annotation improves the recall at the cost of an obvious drop on the precision, which is caused by the significant increase of false positive.

\begin{table}[tbp]
\center
  \caption{Comparison between ${Model}_\text{cls}$ and ${Model}_\text{pred+cls}$ with modified annotations}
  \label{tb:compred2}
\begin{center}
\begin{tabular}{l|c|c|c|c}
\hline
\qquad Model & Acc. & Prec. & Rec. & F1\\
\hline
${Model}_\text{cls}$ & 98.1\% & 87.7\% & \textbf{99.7\%} & 0.933\\
${Model}_\text{pred+cls}$ & \textbf{98.7\%} & \textbf{92.6\%} & 98.0\% & \textbf{0.952}\\
\hline
\end{tabular}
\end{center}
\end{table}

\subsection{Results}

We present several experimental results in Fig.~\ref{fig:result}. In most scenarios, our model can correctly predict the falling event in advance. However, (d) and (e) show two failure cases caused by different reasons. In (d), many keypoints are undetected by OpenPose due to the invisibility of the person's upper body under that camera perspective. As a consequence, the pose prediction module produces absurd keypoints sequence providing insufficient features for classification. (Recall that vectors with less than 8 detected keypoints will be prejudged as `unknown' before the falls classifier). While in (e), no `precursor' of fall is discovered in the observed frame sequence, which is beyond the ability of the pose prediction module to predict the future falls.
%\begin{figure}[htbp]%
%\centering
%\includegraphics[width=1\linewidth]{figs/fig_8.png}
%\caption{Outdoor obstacle segmentation results of different models. (a) The input image. (b) Ground truth of semantic segmentation. (c) Result of PSPNet. (d) Result of DeepLabv3+. (e) Result of ours.}
%\label{fig:outdoor}
%\end{figure}

%\begin{figure}[htbp]%
%\centering
%\includegraphics[width=1\linewidth]{figs/fig_9.png}
%\caption{Segmentation performance of our method on consecutive frames. (a) The input image. (b) and (c) are segmentation result without and with temporal consistency supervision. (d) and (e) are the obstacle detection result of (b) and (c).}
%\label{fig:cons}
%\end{figure}

%\begin{figure}[tbp]%
%\centering
%\includegraphics[width=1\linewidth]{figs/fig_10.png}
%\caption{Obstacle avoidance results for indoor and outdoor scenarios. (a) image (b) segmentation (c) morphological processing (d) path planning}
%\label{fig:res}
%\end{figure}

\section{Conclusion}\label{sec:concl}

In this work, we propose a model combining a pose prediction module and a falls classifier for the precognition of falls. The pose prediction module generates future keypoints vector sequence based on the observation. Then the falls classifier takes the predicted keypoints vector as input and judges whether it's a fall. Evaluations has proved the superiority of keypoints features for falls classification and the effectiveness of the pose prediction module.

{\small
\bibliographystyle{ieee}
\bibliography{egbib}
}

\end{document}